\documentclass[twoside,11pt]{article}
\usepackage{blindtext, graphicx}
\usepackage{url}

\begin{document}

\title{MARVIN: An Open Machine Learning Corpus and Environment for Automated Machine Learning Primitive Annotation and Execution}

\author{Chris A. Mattmann\\
       Jet Propulsion Laboratory\\
       California Institute of Technology\\
       Pasadena, CA 91109 USA\\
       chris.a.mattmann@jpl.nasa.gov 
       \and
       Sujen Shan \\
        Jet Propulsion Laboratory\\
       California Institute of Technology\\
       Pasadena, CA 91109 USA\\
       sujen.shah@jpl.nasa.gov   
       \and
        Brian Wilson \\
       Jet Propulsion Laboratory\\
       California Institute of Technology\\
       Pasadena, CA 91109 USA\\
       brian.wilson@jpl.nasa.gov 
       }

\maketitle

\begin{abstract}
In this demo paper, we introduce the DARPA D$^3$M program for automatic machine learning (ML) and JPL's MARVIN tool that provides an environment to locate, annotate, and execute machine learning primitives for use in ML pipelines. MARVIN is a web-based application and associated back-end interface written in Python that enables composition of ML pipelines from hundreds of primitives from the world of Scikit-Learn, Keras, DL4J, and other widely used libraries. MARVIN allows for the creation of Docker containers that run on Kubernetes clusters within DARPA to provide an execution environment for automated machine learning. MARVIN currently contains over 400 datasets and challenge problems from a wide array of ML domains including routine classification and regression to advanced video/image classification and remote sensing.
\end{abstract}


\section{Introduction}
The world of machine learning (ML) is exploding and new ways of applying ML for mundane, repetitive, and complex tasks is increasing: from using ML to reply to e-mail messages automatically \cite{kannan2016smart} to self-driving cars and using ML for sustainable mobility \cite{burns2013sustainable} . Today, machine learning requires human capital to perform a variety of tasks including 1) composition and aggregation of machine learning ``primitives'' \cite{mitchell2006discipline}, e.g., kernels of code that {\em clean}, {\em decide}, {\em classify/label}, {\em cluster/organize}, and {\em rank} the input dataset to produce an output; 2) selection of ML models that have already been generated such as from OpenML \cite{vanschoren2014openml}, ModelZoo \cite{jia2015caffe}, UC Irvine ML repository \cite{frank2010uci}, Kaggle \cite{wiki:Kaggle}, etc. or the generation of new models, datasets,  and problems; and 3) development of metrics to evaluate how well machine learning is working on a particular problem. 

\begin{figure}
\begin{center} 
\includegraphics[width=0.6\textwidth]{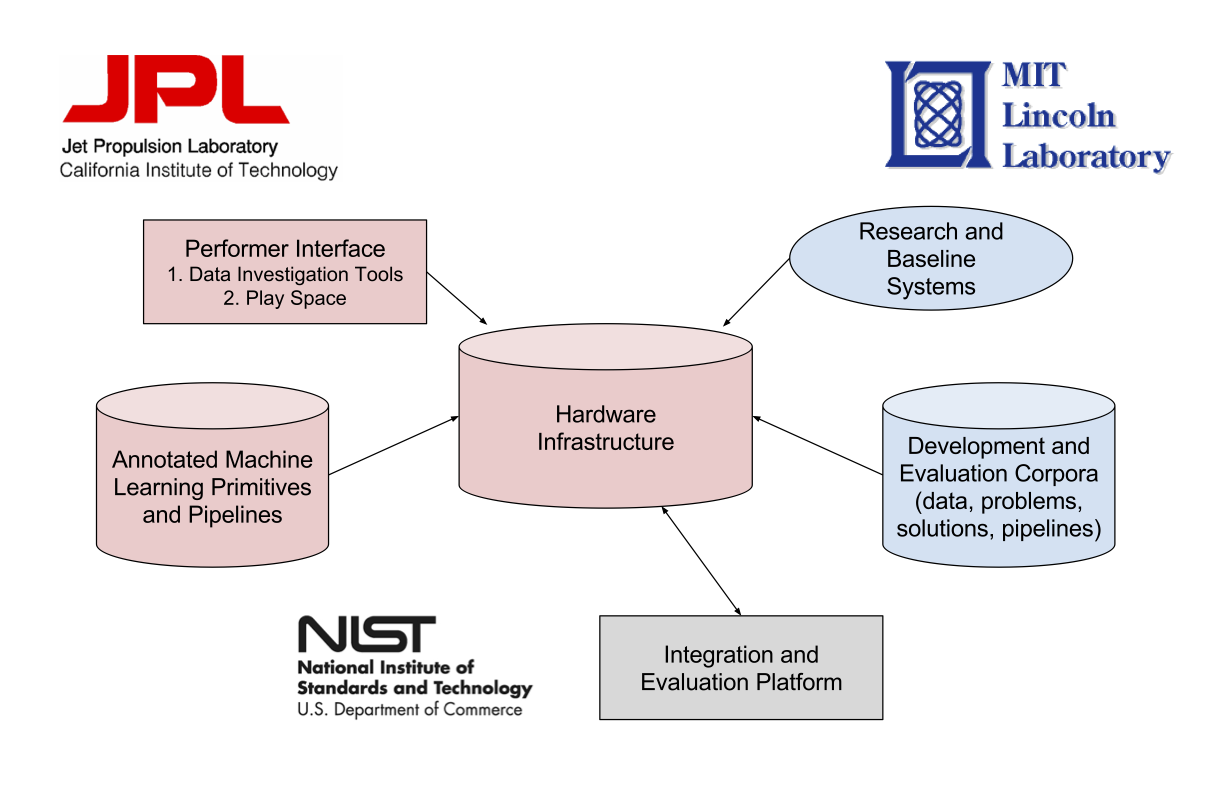}
\end{center} 
\caption{\small \sl D$^3$M Government Team responsibilities.}
\label{fig:1}
\end{figure}

In 2016, the United States Defense Advanced Research Projects Agency (DARPA) incepted the Data Driven Discovery of Models (D$^3$M) program to automate the three core tasks of machine learning identified in the preceding paragraph. Twenty four performers on D$^3$M are working together in three technical areas corresponding to each of the core ML tasks (TA1: ML primitives development; TA2: Automated Machine Learning using Datasets, Problems, and Challenges; and TA3: User/SME Metrics for evaluating automated machine learning). The Jet Propulsion Laboratory, California Institute of Technology (NASA-JPL) is part of the D$^3$M government team, which consists of NASA-JPL, MIT Lincoln Laboratories, and the US National Institutes of Standards and Technology (NIST). The government team responsibilities, shown in Fig.~\ref{fig:1}, involve the creation of a baseline open source model corpus (TA1); the curation of 10s of thousands of ML datasets, problems, and baseline solutions (TA2) for proxy government and even direct governmental challenges across defense, civil agencies, and space; and a set of automated evaluation metrics to assess the performance of the automatic machine learning systems. 

Our team at NASA-JPL has over the past two years designed and implemented a system that we call MARVIN (a reference to the character from the \emph{Looney Tunes} cartoon series \cite{beck2003looney}) whose main goals are to provide an environment for creating a performer baseline for machine learning primitive annotation, schema, and evaluation. MARVIN is a web-based tool that is built on ElasticSearch \cite{gormley2015elasticsearch} and FacetView \cite{facetview2018} and Kibana \cite{gupta2015kibana}. ElasticSearch is used to capture JSON instances of metadata about machine learning primitives representing the world of ML - from SKLearn \cite{pedregosa2011scikit}, to DL4J \cite{bahrampour2016comparative}, to Keras \cite{raschka2015python} and so on. Primitive ``annotations'' capture their inputs, outputs, and characteristics for composing the primitives into an ML pipeline for use by the TA2 performers. MARVIN allows for search and exploration of these primitive instances, and we have written an application programming interface to leverage MARVIN for invocation of the associated primitives (e.g., pick your LogisticRegression that supports particular hyperparameters, programming languages, etc.). In addition, Marvin also provides the ability to search ML datasets, and challenge problems for over 400 challenges at present (and soon 10s of thousands) ranging from introductory machine learning activities and data types e.g., predict Hall of Fame decisions based on baseball player statistics all the way to complex multi-source classification and decision making e.g., identify objects in videos and images with high accuracy.  All primitives, datasets, and challenge problems are searchable and indexed in ElasticSearch and browseable using FacetView and Kibana. 

MARVIN also provides the ability to generate a Dockerfile \cite{} and allow for execution of ML primitives and pipelines based on the primitive type and associated preferred data and problem domain. We have classified domains into areas of video, text/NLP, classification, labeling, regression, etc. MARVIN also provides base Docker images containing the curated library of ML primitives and a base execution environment that D$^3$M performers can extend and use to run generated pipelines. 

\section{Data model}

The Primitive ``annotations'' follow a \cite{primitiveSchema} that allow TA2 systems to infer metadata about the primitive, like its ``algorithm type'', ``primitive family'', ``hyperparameters'', etc. (JPL developed an initial schema but now it is jointly developed by the entire D$^3$M community -- Performers and the government teams.) 

\textbf{Algorithm Types:}
The algorithm type field intends to capture the underlying implementation of the primitive. The field currently is an enumeration of values from a controlled vocabulary including, but not limited to, algorithms such as ADABOOST \cite{adaboost1997}, BAYESIAN LINEAR REGRESSION \cite{minka1998bayesian}, DECISION TREE \cite{decisiontree1986}, etc.

\begin{figure}
\begin{center} 
\includegraphics[width=0.8\textwidth]{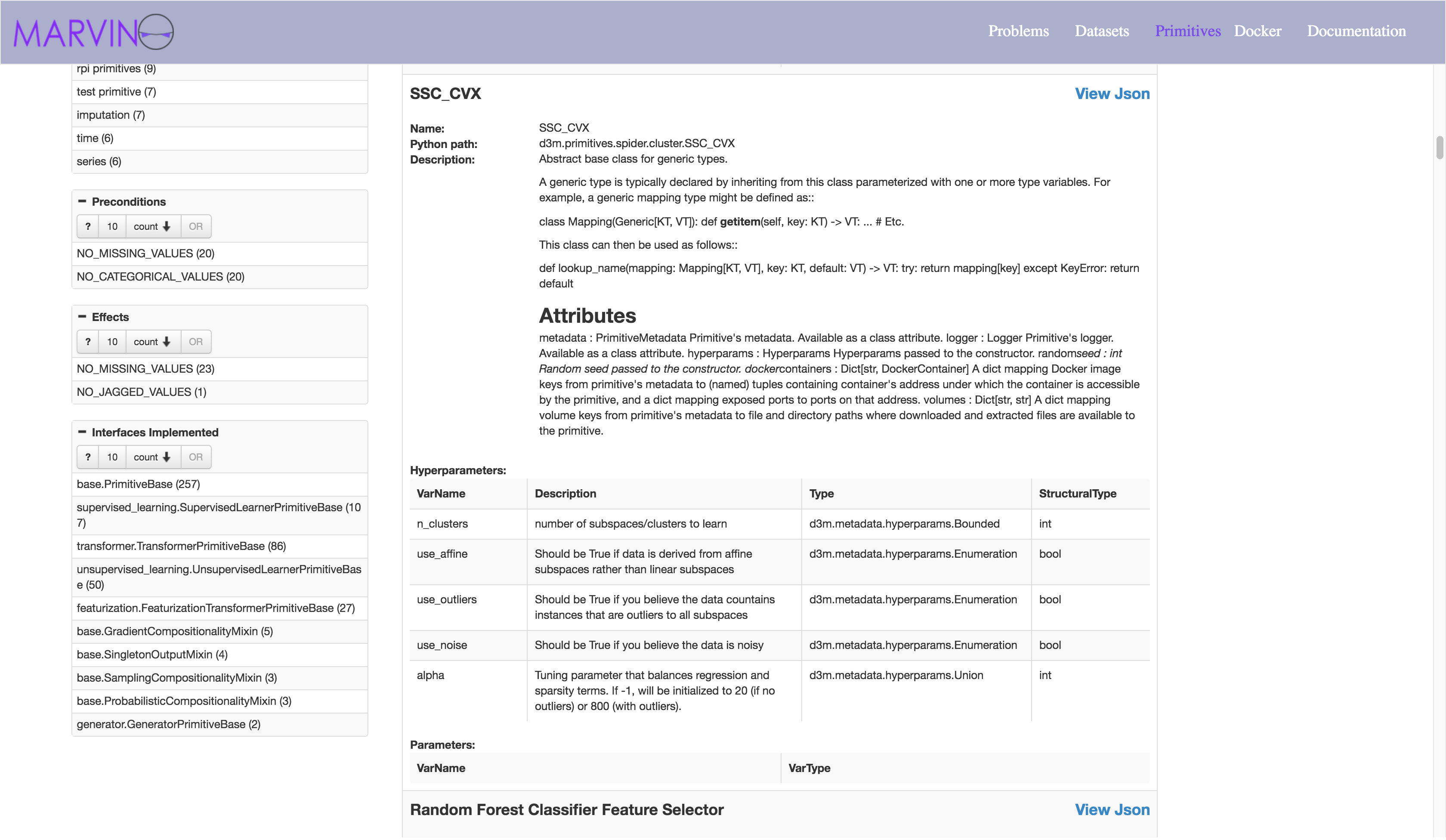}
\end{center} 
\caption{\small \sl A screenshot of the MARVIN interface showing primitives (on the right) and their metadata, along with search criteria on the left, e.g., NO\_JAGGED\_VALUES, etc.}
\label{fig:2}
\end{figure}

\textbf{Primitive Family:}
This field intends to capture the high level purpose of the primitive. It is an enumeration of controlled values e.g., ``CLASSIFICATION'', ``FEATURE\_SELECTION'',
``DATA\_TRANSFORMATION'', ``TIMESERIES\_FORECASTING'', etc.

\textbf{Hyperparameter:} The hyperparameter field captures all the parameters that a TA2 system could pass to a TA1 primitive to control its behavior. These could include tunable hyperparameters which affect the model performance, resource use hyperparameters that control the resource usage of that primitive, and metafeature parameters that control which meta-features are computed by the primitive. 

Apart from the above, there are more key fields that help a TA2 system to determine whether to use the given primitive in a pipeline for solving a problem or not. For example, the \textbf{preconditions} field indicates what preconditions must evaluate to true before using this primitive. For instance the ``NO\_MISSING\_VALUES'' precondition would inform a TA2 system that this primitive cannot handle missing values in the data. Similarly there could be other preconditions like ``NO\_CONTINUOUS\_VALUES'', ``NO\_JAGGED\_VALUES'', etc., as show in Fig.~\ref{fig:2}.

The \textbf{effects} field in the schema informs a TA2 about the set of post conditions obtained by data processed by this primitive. Some of the effects include - ``NO\_MISSING\_VALUES'', this indicates that the primitive removes missing values, ``NO\_CATEGORICAL\_VALUES'', indicates that the primitive removes categorical columns, etc.

\section{Execution environment}
\begin{figure}
\begin{center} 
\includegraphics[width=0.8\textwidth]{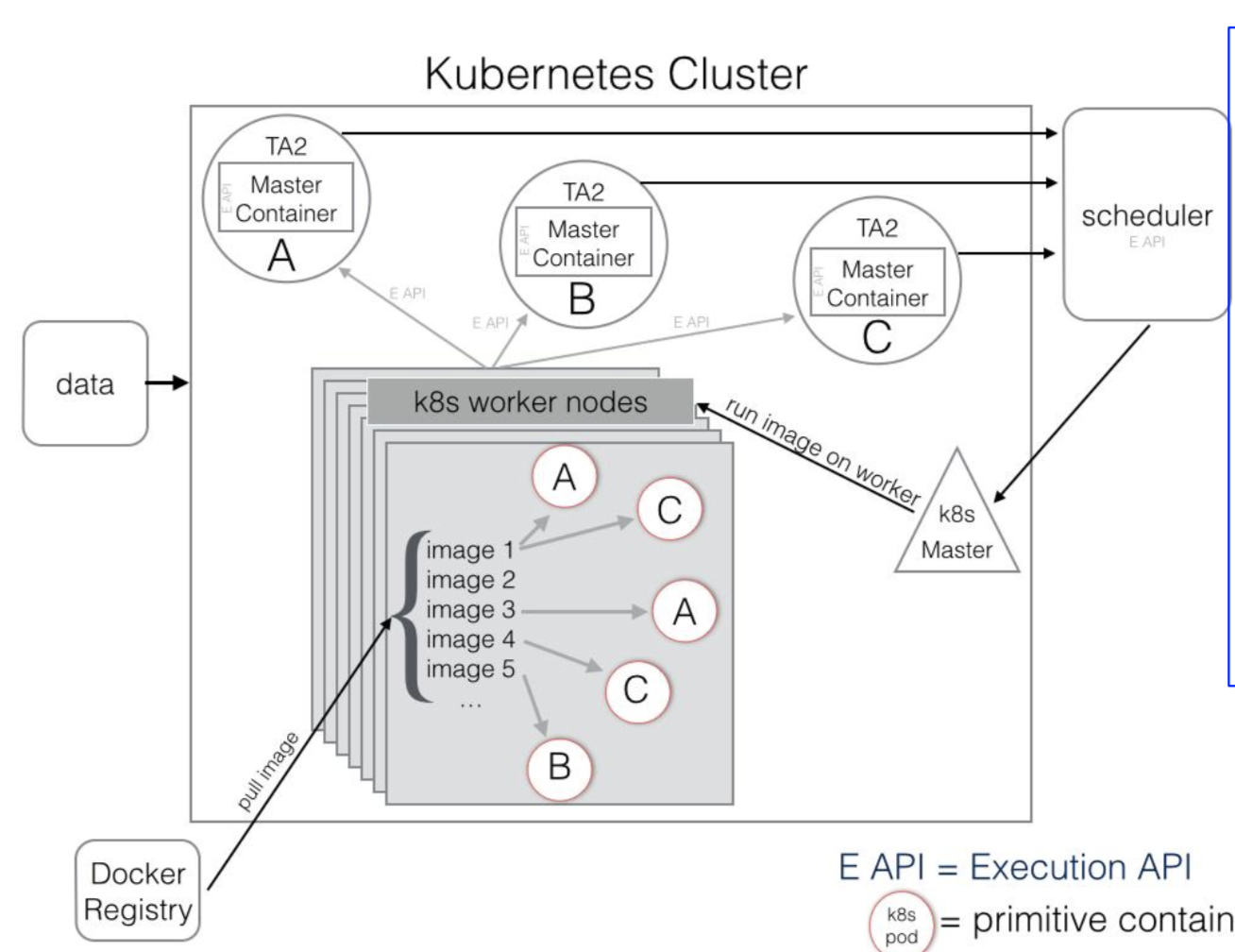}
\end{center} 
\caption{\small \sl Our MARVIN execution environment architecture running on top of Kubernetes (K8s).}
\label{fig:3}
\end{figure}
MARVIN leverages Docker containers and Kubernetes clusters (K8s) \cite{bernstein2014containers} to provide an environment for D$^3$M performer testing (called the ``Play Space'' in Fig.~\ref{fig:1} upper left corner).  Depending on the machine learning problem type e.g., Image/Video, NLP/Text, Classification, performers can use MARVIN to create a Dockerfile containing a set of Primitives for a particular problem set and then create a running Docker image on our D$^3$M Kubernetes cluster. As shown in Fig.~\ref{fig:3} starting from the left side of the figure, TA1 performers can publish a Docker image of their TA1 primitive if it requires special services such as e.g., graph databases, or particular Python libraries which can then be consumable and included in other performer containers. In addition, we offer base Master containers for our TA2 systems (top-middle of Fig.~\ref{fig:3}) to leverage e.g., NLP, Image/Video base, or a combination of all required libraries in the program. Then, using MARVIN's API, TA2 performers can dynamically select, and execute their primitives. Data (shown in the middle-left of Fig.~\ref{fig:3}) is made available via shared K8s volumes to all the containers on the cluster.

\section{Conclusion}
We presented MARVIN, an automated system for discovering machine learning primitives, and executing them in an automated fashion. MARVIN is a core tool in the DARPA D3M program. At present MARVIN provides a web-based interface to browse over hundreds of ML primitives and 400 ML datasets and challenge problems, and over the next two years the tool will be expanded to support thousands of ML problems and primitives for the performers in the program.


\section*{Acknowledgement}
This effort was supported in part by JPL, managed by the California Institute of Technology on behalf of NASA, and additionally in part by the DARPA Memex/XDATA/D3M/ASED programs and NSF award numbers ICER-1639753, PLR-1348450 and PLR-144562 funded a portion of the work. The authors would like to thank Maziyar Boustani, Kyle Hundman, Asitang Mishra, Tom Barber and Giuseppe Totaro at JPL who materially contributed to the design and construction of MARVIN. We also thank our collaborators at MIT Lincoln Labs and NIST. The library of primitives is now being further populated by the entire D3M community. We also credit Mr. Wade Shen for his support of our work.


\bibliographystyle{IEEEtran}
\bibliography{sample}

\end{document}